\documentclass[lettersize,journal]{IEEEtran}

\usepackage{amsmath,amsfonts}
\usepackage{algorithmic}
\usepackage{enumitem}
\usepackage{algorithm}
\usepackage{array}
\usepackage[caption=false,font=normalsize,labelfont=sf,textfont=sf]{subfig}
\usepackage{multirow}
\usepackage{amsthm}
\usepackage[normalem]{ulem}
\newtheoremstyle{Remark} 
{3pt} 
{3pt} 
{} 
{} 
{\bfseries} 
{:} 
{.5em} 
{} 

\theoremstyle{Remark}

\usepackage{textcomp}
\usepackage{stfloats}
\usepackage{url}

\usepackage[colorlinks,urlcolor=blue,linkcolor=blue,citecolor=blue]{hyperref}
\usepackage{verbatim}
\usepackage{graphicx}
\usepackage{cite}
\usepackage[flushleft]{threeparttable}

\usepackage{xcolor}

\newcommand{\revised}[1]{\textcolor{black}{#1}}
\newcommand{\rev}[1]{\textcolor{black}{#1}}
\newcommand{\re}[1]{\textcolor{black}{#1}}
\newcommand{\revf}[1]{\textcolor{black}{#1}}

\begin{document}

\title{Tactile Robotics: An Outlook}

\author{Shan Luo, \textit{Senior Member}, \textit{IEEE}, Nathan F. Lepora, \textit{Member}, \textit{IEEE}, Wenzhen Yuan, \textit{Member}, \textit{IEEE}, \\ Kaspar Althoefer, \textit{Senior Member}, \textit{IEEE}, Gordon Cheng, \textit{Fellow}, \textit{IEEE},
and Ravinder Dahiya, \textit{Fellow}, \textit{IEEE} 

\thanks{Shan Luo is with the Department of Engineering, King's College London. E-mail: shan.luo@kcl.ac.uk.}
\thanks{Nathan Lepora is with the School of Engineering Mathematics and Bristol Robotics Laboratory, University of Bristol. E-mail: n.lepora@bristol.ac.uk.}
\thanks{Wenzhen Yuan is with the Department of Computer Science, University of Illinois Urbana-Champaign. E-mail: yuanwz@illinois.edu.}
\thanks{Kaspar Althoefer is with the School of Engineering and Materials Science, Queen Mary University of London. E-mail: k.althoefer@qmul.ac.uk.}
\thanks{Gordon Cheng is with the Institute for Cognitive Systems (ICS), Technische Universität München. E-mail: gordon@tum.de.}
\thanks{Ravinder Dahiya is with the Department of Electrical and Computer Engineering,
Northeastern University. E-mail: r.dahiya@northeastern.edu.}
}

\markboth{IEEE Transactions on Robotics, Vol. 00, No. 0, Month 2025}%
{Luo \MakeLowercase{\textit{et al.}}: Tactile Robotics: An Outlook}


\maketitle

\begin{abstract}

\re{Robotics research has long sought to give robots the ability to perceive the physical world through touch in an analogous manner to many biological systems.} \re{Developing such tactile capabilities is important for numerous emerging applications that require robots to co-exist and interact closely with humans.} \re{Consequently, there has been growing interest in tactile sensing, leading to the development of various technologies, including piezoresistive and piezoelectric sensors, capacitive sensors, magnetic sensors, and optical tactile sensors. These diverse approaches utilise different transduction methods and materials to equip robots with distributed sensing capabilities, enabling more effective physical interactions.} These advances have been supported in recent years by simulation tools that generate large-scale tactile datasets to support sensor designs and \re{algorithms to interpret and improve the utility of tactile data}. The integration of tactile sensing with other modalities, such as vision, as well as with action strategies for active \re{tactile perception highlights} the growing scope of this field. \re{To further the transformative progress in tactile robotics, a holistic approach is essential. In this outlook article, we examine several challenges associated with the current state of the art in tactile robotics and explore potential solutions to inspire innovations across multiple domains, including manufacturing, healthcare, recycling and agriculture.}
\end{abstract}


\section{Introduction}

\IEEEPARstart{T}{actile} robotics has emerged as a key frontier in robotics research holding considerable potential for transformative advances in robotic applications. The ability to imbue robots with a tactile sense not only enables them to mimic human capabilities, but also signifies a paradigm shift in the ability of robotic systems to perceive and dexterously interact with their surroundings. Tactile sensing 
empowers interactive robots to perform tasks that demand dexterity, adaptability, and understanding of physical interactions~\cite{dahiya2022sensory}.

Over the past few decades, \rev{numerous tactile sensors have been developed} using diverse sensing principles, a broad range of materials and various fabrication methods. Alongside these developments, there have been many attempts to integrate sensors into robotic manipulators and across entire robotic entities, so as to provide them with tactile capabilities. By combining tactile data with other sensing modalities, researchers hope to imbue robots with multi-modal sensing capabilities which would lead to robust perception of the surroundings. The diversity of tactile sensors has given rise to a spectrum of data types, encompassing vibration waveforms, arrays of pressure values, images (or video sequences) and spiking events. Each data type necessitates distinct methodologies to extract meaningful preliminary information, \revised{such as force, contact location, pressure distribution and texture.} \rev{Extracting this information from tactile data and leveraging it to enhance a robot's ability to interact with natural and human-created worlds is the fundamental challenge of tactile robotics}. 




Over a decade ago, Dahiya et al.~\cite{dahiya2009tactile} outlined the research in tactile sensing from humans to humanoid robots. In recent years, there have been several surveys on tactile sensing in robotics, each with its own specific focus. Yousef et al.~\cite{yousef2011tactile} and Kappassov et al.~\cite{kappassov2015tactile} focused on tactile sensing in dexterous robot hands. The review paper by Luo et al.~\cite{luo2017robotic} addressed the robotic tactile perception of object properties. More recently, Li et al.~\cite{li2020review} presented a hierarchy of tactile information and its application in robotics, while the works of \cite{li2024vision} focused on optical tactile sensing -- an emerging methodology. The authors of~\cite{roberts2021soft} presented the state of the art on tactile sensing skins for robots. 
There are also surveys of specific applications of tactile sensing in robotics such as active tactile perception~\cite{seminara2019active}, rehabilitation~\cite{ozioko2022smart} and \revised{minimally invasive surgery (MIS)}~\cite{bandari2019tactile}.

\begin{figure*}[t]
\centering
    \includegraphics[width=0.9\linewidth]{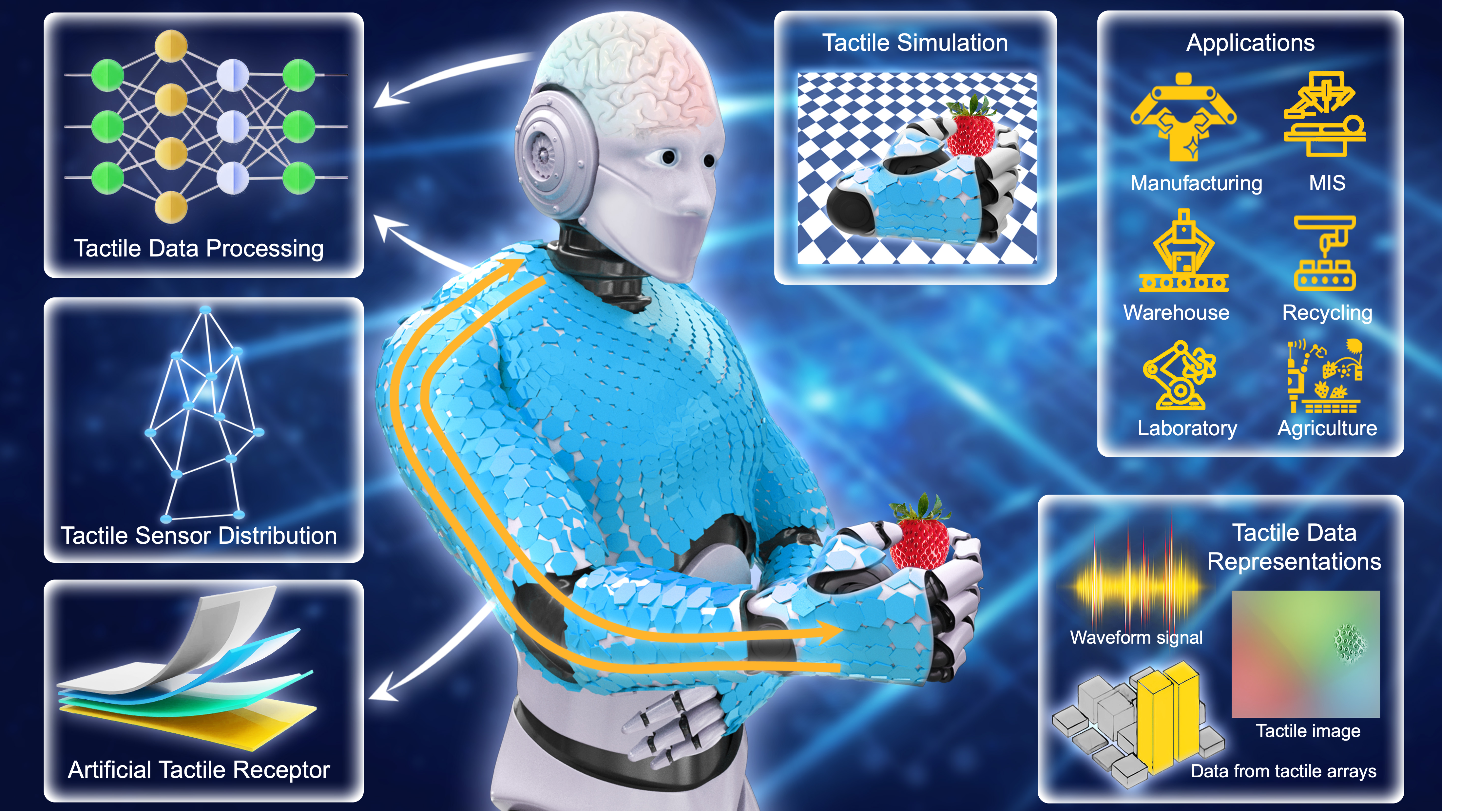}
        \vspace{-.5em}
   \caption{\revf{An overview of the future landscape of tactile robotics, highlighting key advancements across sensing, processing, and simulation. Innovations in materials and transducers will enhance tactile data collection, including waveform signals, data from tactile arrays (e.g., capacitive~\cite{Schmitz2010tactile} or magnetic~\cite{tomo2017covering} arrays), and tactile images from vision-based tactile sensors. Tactile sensors are expected to expand beyond fingertips to cover the entire robot body, forming a dense tactile sensory network. This network will support both local and centralised processing, allowing robots to interpret contact information and adapt their actions accordingly. Tactile simulation is anticipated to play a growing role, supporting new sensor designs and enabling predictive planning.  Collectively, these developments will advance robots toward human-like tactile dexterity with broad application potential.} }
 \label{fig:outlook}
     \vspace{-1em}
\end{figure*}

\re{These survey papers provide general progress reports on specific niche areas of tactile sensing. In contrast, this article aims to define the evolving landscape of tactile robotics as an emerging research field. We review recent progress in the field and anticipate key developments over the next decade (Fig.~\ref{fig:outlook}).} Specifically, we revisit the research areas discussed in the influential paper by Dahiya et al.~\cite{dahiya2009tactile} that provided an early review of tactile sensing, covering tactile sensor types, distributed tactile sensing, and the system-level perspective on tactile sensing. Building on this foundation, we expand the scope to include recent developments, such as simulation tools for tactile sensing, benchmarking, and tactile data interpretation. We also highlight emerging topics such as multimodal perception, which integrates tactile sensing with other modalities like vision, and more traditional topics such as active touch as critical directions for future research.




This paper is structured as follows: 
we introduce the key preliminary concepts of tactile robotics in Sec.~\ref{sec:preliminaries}. Tactile materials, the broad categories of sensors, and  \revised{tactile sensor distribution} are presented in Sec.~\ref{sec:skins} through Sec.~\ref{sec:network}. The tactile simulation is discussed in Sec.~\ref{sec:simulation}. We then detail our insights into tactile data collection and benchmarking in Sec.~\ref{sec:benchmarking}, data interpretation in Sec.~\ref{sec:perception}, multi/cross-modal perception in Sec.~\ref{sec:multicross} and active touch in Sec.~\ref{sec:active}. We conclude the paper in Sec.~\ref{sec:conclusion} with a discussion of future tactile robotics.

\section{Preliminaries}
\label{sec:preliminaries}

Although \revised{\textbf{tactile robotics}} is an emerging sub-field in robotics, there appears, despite an extensive search, to be no firm definition of it within the literature. In this article, we define \revised{tactile robotics} as a field of robotics that focuses on the development and integration of \revised{tactile-sensing technologies} into robotic systems. 
The goal of \revised{tactile robotics} is to enhance a robot's ability to perceive and interact with its surroundings by providing it with a sense of touch.

\textbf{Tactile sensing vs force/torque sensing.} 
We noted a mixture of the use of force/torque sensing with tactile sensing in the literature. \rev{While they share the common goal of addressing touch or contact-based sensing, it is more widely accepted that tactile sensing should involve a spatial distribution of contact readings. In contrast, force/torque sensing provides a single value for each axis, which can be components of a tactile sensing system. Hence, force/torque sensors are often referred to as ``point sensors'', with that point being the location where the torques are measured given the forces.} Research in the wider field of tactile sensing has shown that single force/torque sensors placed at the robot's wrist~\cite{dahiya2009tactile} or fingertip~\cite{liu2015finger} can be used to measure the contact location across the surface of the known shape of the end-effector, using an appropriate modeling strategy. \rev{Beyond most currently available tactile sensors, which are usually limited to measuring normal and lateral forces, this approach can also measure local rotary forces along the axis perpendicular to the contact surface if the used sensor is a six-axis force/torque sensor~\cite{liu2015finger}.} \rev{On the downside, this approach can only determine location and forces when a single contact occurs. Furthermore, current 6-axis force/torque sensors are bulky, expensive, rigid, and prone to damage from overloading, limiting their suitability for integration with robot fingertips.} 

\textbf{Tactile sensing vs haptics/haptic sensing.} Traditionally, the terms ``haptics'' or ``haptic sensing'' have often been intertwined with tactile sensing in the literature, given their shared connection to the sense of touch. Nevertheless, in a broader context, haptics primarily pertains to the feedback conveyed to humans~\cite{hannaford2016haptics}. In tactile robotics, there is a prevalent acknowledgement that haptic sensing encompasses not only tactile sensing at the contact area but also kinesthetic sensing, involving an awareness of the robot's body position~\cite{dahiya2009tactile}. 


Several other new terms have emerged within tactile robotics. In array-based tactile sensors, the individual sensing units are commonly referred to as ``taxels'' (tactile pixels) for tactile elements, analogous to pixels as elemental units of images. These taxels collectively comprise sensing elements for tactile skin, which measures contact parameters distributed over an area. The resulting data format is often represented as an image, giving the term ``tactile image'' for data collected from tactile array sensors or optical tactile sensors~\cite{pezzementi2011tactile,luo2015novel}.

\section{\revised{Materials for Tactile Sensing}}
\label{sec:skins}

 \re{The outermost layer of many biological systems, such as mammals, is the skin. It contains various receptors (sensory elements) that detect changes in the skin's state, such as deformation, stretch, and temperature variation. These changes can result from external contact, environmental factors like wind or sunlight, or internal movements such as joint motion.} Simultaneously, the skin acts as a protective barrier for the underlying receptors and nerves, while possessing the remarkable ability to repair tissue damage. Inspired by these properties, researchers have long sought to equip robots with human skin-like properties by utilising soft materials embedded with sensors distributed at varying densities across the robot's body.




\subsection{Robotic Skin Material}
\label{sec:robotskin}
\rev{Human skin is a flexible, waterproof, self-healing protective covering embedded with receptors for pressure, temperature, pain, and tissue damage~\cite{klatzky_touch_2003}.} 
Its three principal layers, {\em i.e.}, the epidermis, dermis and hypodermis, work together to enable tactile perception~\cite{klatzky_touch_2003,loomis_tactual_1986}. The epidermis is a thin protective layer (75-150 $\rm{\mu m}$), the dermis beneath it provides structural support (1 to 4 mm thick) and the hypodermis comprises variable thicknesses of fatty connective tissue~\cite{goldsmith1991physiology}. At the epidermal-dermal interface, a mesh of dermal papillae protrudes into the epidermis, near which the shallow mechanorceptors are located.    
\rev{This layered structure, coupled with the diverse tactile receptors of the skin, facilitates the perception of fine differences in texture, temperature, and stiffness~\cite{abraira_sensory_2013}.} 

Drawing inspiration from these properties, roboticists have sought to emulate the tactile capabilities of human skin using soft artificial materials. Early work investigated requirements such as elasticity, adhesion, and tactile sensor compatibility, testing materials such as Poron, PDS, Sensoflex and rubber~\cite{cutkosky1987skin}. 
Advances have since expanded to include fabric-based resistive materials, which integrate conductive yarns or coatings to form grid-like structures. These materials detect resistance changes caused by pressure or deformation, making them lightweight, stretchable, and cost-effective for wearable and robotic applications~\cite{cherenack2012smart}. Their flexibility and breathability allow for conformation to complex surfaces and are suitable for human-machine interfaces and prosthetics.  

Another research strand focuses on electronic skins, also \rev{called e-skins}, designed to mimic the functionalities of human skin. Such systems are made of thin, flexible and stretchable layers of conductive materials, flexible electronics, \rev{and} integrated sensors ({\em e.g.}, resistive, capacitive, piezoelectric) to measure pressure, temperature and humidity~\cite{someya2016rise}. 
E-skins often aim to replicate the sensory and mechanical properties of human skin, thus playing a pivotal role in enhancing human-machine interaction and robotic systems~\cite{soni2020soft}.


\rev{Tomography-based tactile sensing, such as electrical impedance tomography (EIT)~\cite{park2021deep} or electrical resistance tomography (ERT)~\cite{zheng2023adaptive}, has also been explored. These methods reconstruct internal conductivity distributions of a given object by applying small electrical currents through surface electrodes and measuring changes in the electrical potential across the object caused by impedance variations within the object. Originally used in fields like medical diagnostics and geophysics, EIT is well-suited for tactile sensing. By introducing currents into a thin resistive layer, external forces alter current paths, allowing estimation of force and location through data-driven calibration methods~\cite{zheng2023adaptive}.} 

\revised {While many tactile sensing solutions incorporate sensing capabilities directly into the outer skin, other techniques use the outer skin primarily as an interface for physical interaction with the environment. In these cases, the outer layer is typically composed of an elastic, deformable material, with its deformations being detected by sensors positioned below it.}
\revised{For tactile sensors that directly measure surface deformation, such as optical tactile sensors, softer skin enhances sensitivity to small contact forces by increasing surface deformation. In contrast, for sensors like barometric sensors or capacitive arrays, in which the sensing elements are covered by an elastomer layer, the relationship is more complex. Softer skin can lead to a broader distribution of load over the sensing elements, which may reduce resolution when detecting small, localised forces and attenuate higher-frequency signals. Therefore, skin softness must be carefully optimised for the specific application requirements.} 

Commonly used materials include thermoplastic elastomers (TPEs) 
and silicone rubbers. 
\revised{Typical Shore values of elastomers range between 5A and 20A for GelSight sensors~\cite{yuan2017gelsight}, 26-28A for the thin outer 3D-printed skin of the TacTip sensor (which covers an inner gel that can range from very soft to stiff)~\cite{ward2018tactip,lepora2021soft}, and 00-50 for uSkin sensors~\cite{tomo2017covering}.} Most optical tactile sensors ~\cite{yuan2017gelsight,gomes2020geltip,ward2018tactip,li2020f} require optically clear elastomer bases to facilitate LED light penetration for deformation capture by the camera. \rev{Instead of clear elastomer, the soft thumb-sized sensor Insight~\cite{sun2022soft} and its successor Minsight~\cite{andrussow2023minsight} require a soft material that is opaque and shiny.} 


\rev{Similar to the stratified layers found in human skin~\cite{klatzky_touch_2003,loomis_tactual_1986}, robotic skins can also benefit from a multilayered design.} For example,  GelSight sensors~\cite{yuan2017gelsight} use a thin silicone layer mixed with reflective pigments above a transparent elastomer to enhance light reflection. 
Beyond encapsulation, materials such as Au, Cu, silver nanowires, Pt, liquid metal, carbon nanotubes (CNTs), ionic materials, and nanofibres are commonly used in electrode and sensing layers~\cite{luo2023technology,pyo2021recent}.

\subsection{Challenges and outlook}
Envisioning the future of robotic materials involves the development of a multilayered composite material characterised by non-homogeneity, non-linearity, visco-elasticity, and anisotropy. This material will need to meet various requirements, in terms of friction, adhesion, resilience, elasticity, durability, 
and suitability for tactile sensing. Achieving this goal involves challenging technical considerations, such as engineering visco-elastic material and utilising advanced manufacturing techniques like 3D-printing for precise control over structure and composition, and laser micromachining, \revised{lithography and electrohydrodynamic (EHD)} printing to produce bio-inspired materials at micro-/nano-scale. Dedicated 3D-printing facilitates the incorporation of sensors for enhanced tactile sensing within complex, non-homogeneous patterns~\cite{nassar2023fully}.

\textbf{Need for a full analysis of tactile skin materials.} \rev{The current landscape of material analysis within tactile sensor skins exhibits a notable fragmentation~\cite{pyo2021recent}. Although a wide range of materials have been explored, as detailed in Section~\ref{sec:robotskin}, each has inherent limitations. For example, elastomers used in~\cite{yuan2017gelsight}, although soft and sensitive, often lack durability under prolonged use or under harsh conditions. Similarly, conductive polymers can suffer from hysteresis~\cite{someya2016rise}, which reduces the accuracy of long-term sensing. Temperature sensitivity is another challenge, as some materials experience performance degradation in dynamic or extreme environments~\cite{luo2023technology}. The lack of a standardised framework for evaluating and comparing these materials hinders a clear understanding of trade-offs between properties such as sensitivity, flexibility, cost, and robustness. To address these challenges, a more systematic approach is needed to benchmark existing materials, explore hybrid formulations, and develop novel materials that overcome current limitations.}

\textbf{\revised{Automating fabrication is key.}} \rev{Many optical tactile sensors rely on manual processes such as mixing, painting, or applying coating, introducing variability that affects consistency~\cite{yuan2017gelsight,gomes2020geltip}. 
In addition, fabric-based sensors often involve handcrafting~\cite{cherenack2012smart} and e-skin technologies typically require numerous manual steps in cleanroom environments~\cite{someya2016rise}. Automated fabrication can address these challenges by improving precision, reducing variability, and ensuring a consistent composition of the tactile sensor components. Future developments of these technologies are crucial to improve the reliability and reproducibility of tactile sensor applications, by ensuring scalability and consistency in production.}

\rev{A comparison of recent research highlights several leading materials for tactile sensors. A recent review~\cite{wang2023tactile} identifies PDMS, PVDF, PZT and hydrogel materials as materials of interest for use in tactile sensors in both robotic and wearable applications. PDMS, a soft elastomer, is widely used for its biocompatibility and ability to form compliant layers, although it is not particularly robust under high strain. PVDF, a piezoelectric polymer, offers flexibility and is well suited for sensing applications that require deformation-based feedback. PZT, in contrast, is a piezoelectric ceramic valued for its high sensitivity and electromechanical coupling, but its rigidity makes it less suitable for applications where a soft physical interaction with the environment is required. Although these materials differ significantly in their mechanical properties, their biocompatibility enhances their utility for applications involving direct human interaction.} Exploring alternative solutions, such as modular designs with replaceable skin layers, could further mitigate these challenges. Examples include the early version of a tactile sleeve developed for the STIFF-FLOP manipulator~\cite{sareh2014bio}. 

\section{Transduction Methods for Tactile Sensing}
\label{sec:transduction}

In tactile robotics, transduction methods play a pivotal role, analogous to the conversion of mechano-received stimuli into sensations in human perception. 
Although there are some survey papers that focus on transduction methods in tactile sensing~\cite{yousef2011tactile,dahiya2009tactile,roberts2021soft}, here we provide a concise introduction to current approaches while adopting a forward-looking perspective that highlights future methods and challenges.

\subsection{Current transduction methods for tactile robotics}
In the human tactile sensory system,  transduction plays a crucial role by converting external stimuli, such as pressure and temperature, into electrical signals within the nervous system. 
Notably, the skin exhibits pyroelectric responses to rapid changes in temperature, and conversely, piezoelectric responses to a pressure pulse~\cite{lumpkin2007mechanisms}. In tactile robotics, various transduction methods have been explored. 



\subsubsection{Piezoresistive and piezoelectric sensors} 

Piezoresistive sensors detect mechanical stress by measuring changes in material resistance~\cite{fiorillo2018theory,hannah2018multifunctional}. \rev{Piezoelectric sensors generate an electric charge in response to stress, offering flexibility, high accuracy, sensitivity, and rapid frequency response~\cite{kim2014dome,zhang2022finger}.} One such example is NeuTouch~\cite{taunyazov2020event}, which uses a layer of electrodes with 39 taxels and a graphene-based piezoresistive thin film for tactile sensing. Its extended version, NUskin~\cite{taunyazov2021extended}, offers a high sampling rate of 4 kHz. \rev{However, piezoelectric sensors are typically limited to dynamic force detection due to voltage decay under sustained pressure~\cite{wang2023tactile}. They also exhibit temperature sensitivity, which affects the measurement stability and accuracy, and can suffer from hysteresis effects.} 

\subsubsection{Capacitive sensors} Applied force or pressure can be measured from the effect on capacitive structures built into a sensor. Advanced micromachining techniques enable the integration of capacitive sensors into intricate force-sensing arrays, as seen, for example, in the \revised{tactile sensors from Pressure Profile Systems}\footnote{https://pressureprofile.com/sensor-systems/sensors}. Multiple capacitors can be embedded into flexible polymers to create capacitive arrays that have been widely used in large-area tactile skins~\cite{schmitz2011methods,maiolino2013flexible,dahiya2019large,liu2022neuro,liu2022printed,mittendorfer2011humanoid,dahiya2013Directions}. They have also been installed into various manipulators, including iCub hands~\cite{schmitz2011methods}, the Allegro hand~\cite{thomasson2022going}, and PR2 grippers~\cite{romano2011human}. 3D-printed hands and palms/soles with intrinsic capacitive sensors have also been reported~\cite{ntagios20223D}. However, capacitive sensors are susceptible to electromagnetic and crosstalk noise and suffer from issues such as hysteresis~\cite{kappassov2015tactile}.


\subsubsection{Magnetic sensors} Magnetic sensors use magnetoresistive or hall-effect sensor chips to measure electrical signal responses to an applied magnetic field. These sensors incorporate magnets within a deformable medium, serving as both sensor skin and magnet enclosure. External forces deform the skin, displacing the magnets and altering the local magnetic fields, and thereby enabling detection by the magnetic sensors. \revised{Examples of magnetic tactile sensors are uSkin~\cite{jamone2015highly}, and ReSkin \cite{bhirangi2021reskin}}. Although magnetic tactile sensors offer certain desired features, such as a linear response to shear forces, high sensitivity and a wide dynamic range, external magnetic field interference can critically impact their operation.


\subsubsection{Optical tactile sensors} Optical tactile sensors detect changes in light intensity or gradients transmitted through a transparent medium, such as elastomer gel, to measure contact information, including pressure and slip. Currently, there are two main classes of optical tactile sensors: camera-based and fiber-optic sensors. Camera-based tactile sensors leverage the widespread availability of CCD and CMOS imaging arrays, and have been prevalent in recent years thanks to the ongoing miniaturisation and reduction in cost of this technology. One type uses the graduation in reflected light intensity to capture elastomer surface deformation, with examples including GelSight~\cite{yuan2017gelsight} and its variants such as GelSlim~\cite{donlon2018gelslim}, GelTip~\cite{gomes2020geltip}, \re{DIGIT~\cite{lambeta2020digit}, DIGIT 360~\cite{lambeta2024digit360},} TouchRoller~\cite{cao2023touchroller}, GelFinger~\cite{lin2023gelfinger} and ViTacTip~\cite{fan2024vitactip}. Another type tracks markers attached to the inner sensor surface, such as the family of TacTip sensors~\cite{ward2018tactip} that place markers on papillae-like structures to amplify contact. \revised{Likewise, there are many other examples of marker-based optical tactile sensors, including~\cite{ferrier2000reconstructing}, GelForce~\cite{sato2009finger}, \cite{sferrazza2019design}, ChromoTouch~\cite{lin2019sensing}, F-TOUCH~\cite{li2020f}, and FingerVision~\cite{yamaguchi2017implementing} among others}.

Other camera systems, such as depth cameras ~\cite{huang2019depth}, event-based cameras~\cite{kumagai2019event}, \revised{thermal camera~\cite{abad2021haptitemp}} and multiple cameras~\cite{padmanabha2020omnitact,li2023marker}, have also been used for tactile sensing. \revised{There is also work on combining surface sensing and proximity sensing by using different light sources or surface coating technologies~\cite{shimonomura2016robotic, yin2022multimodal}. }Miniaturisation remains a challenge due to focal length limitations, primarily restricting these sensors to fingertip applications, although \revised{a recent attempt to address this issue involves the use of fiber optic bundles~\cite{di2024using}.} Fiber-optic sensors measure light intensity in fibers to detect tactile properties such as pressure. Their compatibility with \rev{magnetic resonance imaging} (MRI) environments, makes them ideal for {\em in vivo} experiments and clinical trials~\cite{xie2013fiber,dawood2022real}. \revised{Light-based detection devices such as photodetectors~\cite{piacenza2020sensorized} and solar cells~\cite{escobedo2020energy} have also been used for tactile sensing, detecting pressure patterns, object movement and proximity~\cite{chirila2024self}.} Solar cells, in particular, have the advantage of being self-powered.

\subsubsection{Other approaches} In addition to the transduction methods outlined above, a range of other approaches have been explored. Among them are vibration-based tactile sensing using microphones~\cite{yang2021large}, the use of acoustic waveguides~\cite{chossat2020soft}, EIT~\cite{park2021deep}, textile based sensing~\cite{buscher2015flexible}, barometer-based ~\cite{koiva2020barometer}, and ionic tactile sensing~\cite{amoli2020ionic}. Multimodal tactile sensors have also been developed, with the SynTouch BioTac sensor used widely in the 2010s. \re{It measures changes in impedance within a conductive fluid contained below the outer skin, using 19 spatially distributed electrodes. When the elastic skin deforms under contact, it reshapes the underlying fluid volume, leading to measurable impedance variations.} The BioTac sensor also integrates DC and AC pressure, and DC and AC temperature for multimodal sensing~\cite{fishel2012sensing}.


\subsection{Challenges and Outlook}

\textbf{Which transduction approach is better?} \rev{Currently, there is no single transduction method that researchers can agree is better than its rivals, as each comes with trade-offs depending on the specific robotic applications. This is mainly due to the wide variety of key metrics for different robotics tasks, {\em e.g.}, spatial resolution, structural dimension, temporal resolution (frequency), sensing range and sensitivity, compliance and durability of the sensing surface, repeatability and hysteresis. For example, vision-based tactile sensors~({\em e.g.} \cite{yuan2017gelsight,ward2018tactip,gomes2020geltip}), while capable of offering high spatial resolution, create challenges in terms of efficient data processing and real-time performance. In contrast, methods such as capacitive or piezoresistive sensing often provide lower data volumes but may sacrifice spatial resolution or sensitivity~\cite{schmitz2011methods,liu2022printed}. The selection of the optimal approach, therefore, depends on the specific requirements of the task at hand. As the field evolves, hybrid or integrated transduction approaches, such as the incorporation of force-sensitive mechanical structures~\cite{althoefer2023miniaturised} or high-frequency microphones~\cite{pestell2022artificial-2}, show promise in addressing the limitations of individual methods.}

\textbf{Low-cost tactile sensors.} The high fabrication cost of tactile sensors has hindered their widespread adoption. 
\rev{Overcoming cost barriers can accelerate the integration of tactile sensing into various applications, leading to low-cost optical tactile sensors such as the DIGIT~\cite{lambeta2020digit} (retails at \$350) and the GelSight Mini (retails at \$500), which are far cheaper than commercial tactile sensors such as the BioTac (used to retail at \$5,000-10,000). Also, some sensors can be 3D-printed inexpensively, such as the DigiTac~\cite{lepora2022digitac} customisation of the DIGIT, although access to fabrication facilities such as multi-material 3D-printers can act as a barrier for adoption.}

\textbf{Coupling the transduction with software algorithms.} Successful implementation of tactile sensors hinges on concurrent development of both physical hardware and the accompanying algorithms. The synergy between hardware and software is critical to the interpretation of sensor data and therefore ultimately to the creation of commercially viable and fully functional robotic systems. \rev{For instance, convolutional neural networks (CNNs) are now a widely used technique  to process data from image-based tactile sensors that make use of the high spatial resolution for improved feature extraction. \re{Similarly, frequency decomposition techniques are utilised to analyse vibration data~\cite{sinapov2011vibrotactile}, enabling fine-grained detection of surface textures and contact dynamics.} Another example is the use of neuromorphic computing frameworks, as demonstrated in~\cite{wang2023neuromorphic}, which efficiently process tactile information to mimic biological sensory processing. These advancements underscore the importance of co-designing hardware and software to achieve optimal performance in tactile sensing applications.}



\section{Tactile Sensor Networks}
\label{sec:network}

When multiple tactile sensors or different transduction methods are installed in a robot, they need connecting to each other and to the central processing unit; collectively, this form a sensor network, also known as a skin-cell network~\cite{cheng2019comprehensive}. 

\subsection{Current sensor networks for tactile robotics}


\rev{Human skin functions as a remarkable bio-electronic network crucial for tactile perception, acting both as a transmitter and receiver~\cite{saint2023human}. Covering a substantial surface area of 1.5–2~${\rm m^2}$ in adults, the skin forms a web of cells and transmitter channels, resembling a dual-role receiver and transmitter antenna. Its electronic organisation relies on a dense network of myelinated fibers, corpuscles (Pacini, Ruffini, and Meissner), and Merkel cells, serving as efficient sensors for various stimuli encountered by epidermal and dermal cells~\cite{lumpkin2007mechanisms}. These stimuli result in electron-based signals, characterised by very low voltages and intensities (mV, nA), which are transmitted to the brain at speeds ranging from 33 to 75 m/s~\cite{lumpkin2007mechanisms,lederman2009haptic}. Although these skin sensors are distributed across the entire human body, they exhibit a much higher density (by 4 to 5 times) in the fingertips and lips~\cite{lederman1981texture}, which are the crucial areas for tactile exploration and sensory discrimination.}

The development of tactile sensor networks for robots has progressed significantly in recent years, particularly in covering the robot's body with tactile sensors arranged in regular matrices. Early efforts focused on integrating sensors into robot arms for obstacle avoidance~\cite{cheung1989proximity}, then flexible sensor matrices with pressure and temperature sensors were also introduced to create conformable robot skin~\cite{someya2005conformable}. However, these matrices had high wire counts and were not robust. Modular systems have also been explored, equipping robots with tactile sensing modules distributed across various body parts. Examples include the RI-MAN robot which had tactile sensing modules on its arms and chest~\cite{mukai2008development} 
and the ARMAR-III robot which had force sensor matrices on its shoulders and arms~\cite{asfour2006armar}. The TWENDY-ONE robot achieved complete coverage with force sensors distributed on its hands, arms, and trunk~\cite{iwata2009design}. RoboSkin, a modular robot skin system, utilised triangular skin modules with capacitive force sensors~\cite{cannata2008embedded,albini2021exploiting}, offered flexibility and significantly less wiring. These modular systems facilitate hierarchical architectures with real-time communication, enabling effective tactile perception. 

Efficient organisation and transmission of tactile information is crucial for robot skin networks. \revised{Physiological mechanoreceptors exhibit overlapping receptive fields of skin areas to which they are sensitive. This allows information from multiple skin sensors to be combined to achieve high sensing acuity known as tactile hyperacuity~\cite{loomis1979investigation} and represents a design template for artificial skins~\cite{lepora2015tactile,sun2022guiding}.} Furthermore, a query-based modular readout system combined with real-time middleware has been utilised to achieve efficient organisation 
of tactile information~\cite{youssefi2015real}. Neuromorphic principles have also been employed to reduce redundancy at the sensor level, significantly reducing network traffic and computational power requirements~\cite{bartolozzi2017event}. The communication network for robot skin systems must be stable, robust, and provide reliable bi-directional connections between skin cells and the host PC. To address these challenges, a self-organising network protocol for skin cells has been developed~\cite{cheng2019comprehensive}, 
which enables automatic construction of bi-directional communication trees, dynamic online re-routing of connections, and facilitates load balancing to optimise network performance.

\subsection{Challenges and outlook}
Although most of the current tactile sensor networks are homogeneous, consisting of tactile sensors of the same type, a future focus could be to integrate different types of sensors into the same network to meet the mounting needs of a system.  
For example, by using optical tactile sensors~\cite{yuan2017gelsight,ward2018tactip,gomes2020geltip} for the fingertips and capacitive sensors~\cite{dahiya2019large,albini2021exploiting,dawood2020silicone} for the robot arms,  one could offer \revised{complementary} sensing capabilities tailored to specific tasks. Integrating sensors of different types naturally introduces new challenges, particularly when designing interfaces to facilitate communication and data exchange between these sensors. However, these are challenges that must be faced to advance the capabilities and effectiveness of tactile skin networks.

The powering of distributed sensors and associated electronics is another significant challenge that has not received much attention, despite its relevance in autonomous robots. The energy sources in human skin are, in essence, distributed in a similar fashion to touch sensors, and an equivalent approach for tactile skin is needed for stable operation over longer times. The energy distribution method may well impact the design of robotic platforms. For example, there will be no longer be any need for backpack batteries in humanoids. Recent advances in energy generation and storage (\revised{{\em e.g.}, printed batteries, solar cells}) and their potential applications in a variety of robots are discussed in~\cite{mukherjee2023bioinspired}. Among these, the use of miniaturised solar cells as touch sensors is noteworthy, because they examplify multifunctional devices that ease the challenges of integration~\cite{escobedo2020energy}.

\section{Simulation for Tactile Sensing}
\label{sec:simulation}



Simulation is an important tool to verify system design and for low-cost data collection. With increasing demand for the integration of tactile sensing into robotic devices, there is a corresponding increase in demand for tactile sensor simulators.  First and foremost, simulation can address the issue of the high cost of tactile data collection, which is a combined consequence of demand for increasingly larger datasets and tactile sensors being vulnerable to breakages from the necessity of physical contact during data collection. Simulation can also alleviate issues due to the high price and \revised{fragility} of some tactile sensors.

When considering sensor simulation methods, we should consider the working principles that underlie tactile sensors. These sensors quantify contact using a deformable body as the contact medium. This could be a piece of elastomer ({\em e.g.} \cite{yuan2017gelsight,gomes2020geltip,lambeta2020digit}), a combination of elastomer over a gel~({\em e.g.} \cite{ward2018tactip,lepora2022digitac}), a wrapped body of gel, liquid or air~({\em e.g.} \cite{alspach2019soft,yamaguchi2017implementing}), or a porous structure \cite{tomo2018new}. The medium  deforms under contact forces, and the sensors use different transduction methods to translate the deformation into electronic signals, such as using capacitance changes or light with an internal camera (see Sec. \ref{sec:transduction} for details). These processes are complex and often cause differences between the actual sensor’s readings and the theoretical results. Researchers have developed a range of methods to simulate tactile sensors. Most of them focus on simulating one aspect: either modelling the deformation of the medium under external forces, or the transducer’s performance in transferring the deformation into electrical readings. 

In humans, we can simulate tactile sensing by mentally creating and manipulating sensory perceptions related to the sense of touch without any actual physical stimuli; {\em i.e.}, tactile mental imagery or somaesthetic mental imagery~\cite{gallace2013somesthetic}. The process involves generating mental representations of tactile sensations or textures in the absence of direct physical contact. Similar to visual imagery~\cite{pearson2015mental}, individuals can mentally simulate the feeling of different textures, temperatures, or pressures on their skin. Imagining the feel of soft fabric or rough surfaces engages in specific tactile mental imagery. This cognitive process allows individuals to mentally experience and explore tactile sensations, contributing to overall sensory perception. Tactile mental imagery finds applications in rehabilitation, where individuals use it to practice movements or interactions \revised{without actual physical engagement}, and in design, where it informs the development of user-friendly products. Drawing inspiration from human tactile mental imagery, robots can simulate and interact with various tactile stimuli in a virtual environment. In human-robot interaction, robots equipped with tactile simulations can better understand and respond to human touch, leading to safer and more intuitive interactions.


Researchers have integrated tactile simulators with robot simulators, examples being GraspIt!\cite{miller2004graspit}, Gazebo\cite{Koenig-2004-394}, MuJoCo\cite{todorov2012mujoco}, PyBullet\cite{benelot2018}\revised{, and Drake\cite{drake}}. Those simulation engines use simplified contact models, which limit the match that is possible with real tactile sensor behavior; even so, sophisticated applications such as multi-fingered in-hand manipulation have been achieved~\cite{qi2023general,yang2024anyrotate}. A richer interface between tactile and robot simulations is provided through geometry-based methods~\cite{gomes2019gelsight,gomes2021generation,gomes2023beyond,zhao2024fots}. Here, the tactile simulator takes the local geometry of the external object at the contact surface as the input, and simulates the sensor’s raw output. This geometry is mostly sampled using a depth camera placed below the contact surface in the robot simulator~\cite{gomes2021generation,zhao2024fots}. In some cases, contact forces are also simulated~\cite{zhao2024fots}. We next review simulation methods that convert input geometry into tactile sensor output using the three different methods.


\subsection{Current methods for robotic tactile simulation}

\textbf{Physics-based Methods.}
Simulating tactile sensors often involves building physical models to replicate sensor readings. Some studies focus on modeling the deformation of the soft medium under external force and torque, typically using \revised{finite element methods (FEM)}. 
One example is to develop a local contact force model for the sensor surface and generated meshed surfaces for simulation~\cite{moisio2012simulation}.  
Similarly, another example utilised a calibrated FEM model to simulate the deformation of BioTac fingertip-shaped sensors~\cite{narang2021sim}, employing a GPU-based Newton solver to accelerate calculations. However, the calculation cost is the main limitation with these methods. To address this cost, researchers have explored alternative simplified graphical models, such as particle fields, for efficient soft-body deformation calculation~\cite{chen2023tacchi}. Additionally, a simplified mechanical model can be used to describe the motion of the sensor's soft surface to accelerate computation~\cite{xu2023efficient}. 

Other studies focus on simulating the transducers in the sensors. For example, GelSight sensors \cite{yuan2017gelsight} utilise optical reflection and refraction to translate the deformation of the sensing medium into light intensity (see Section~\ref{sec:transduction}). The imaging process within the sensor can be simplified by employing the Phong reflection model~\cite{phong1998illumination} as a simplified rendering model to simulate the reflected light from the sensor's surface~\cite{gomes2019gelsight,gomes2021generation}. This method can be further improved by introducing a geometric path model to approximate the light field for curved optical sensors using light-piping illumination~\cite{gomes2023beyond}. 
\revised{Another optical simulation method is the physically-based rendering~\cite{pharr2023physically} adopted in~\cite{wang2022tacto}. This method synthesises a camera's output of an optical scene by tracking optical rays interacting with the optical components in the scene.} 
However, the performance relies on the precision of the light simulator, with a trade-off between computational efficiency and accuracy, particularly in environments with complex optical designs that cause multiple light ray bounces.

\textbf{Data-driven Methods.}
Data-driven methods for simulating tactile sensors involve using large-scale datasets of sensor readings and machine learning techniques to train models to predict sensor readings in new contact scenarios. The widespread adoption of deep learning has made this approach more practical for modelling the complex behavior of tactile sensors. One example used conditional Generative Adversarial Networks (cGAN) to simulate a GelSight sensor based on data collected from an RGB-D camera~\cite{patel2020deep}. Another example simulated point contact with a vision-based marker tactile sensor using data from FEM models and a \revised{CNN} model~\cite{sferrazza2020learning}. Also, the TacTip sensor can be modelled by collecting a dataset of the sensor's membrane deformation in the Unity graphical software and training a neural network for prediction~\cite{ding2020sim}. 

Another type of data-driven approach relies on using a simplified physics simulation, such as object penetration with no skin dynamics, relying instead on tactile image-to-image translation to bridge a substantive sim-to-real gap. One example simulated the TacTip sensor using GANs for image-to-image translation between geometric output in a robot simulator (a depth map) and readings from a real sensor (marker shears)~\cite{church2022tactile}. Due to its simplicity, robot manipulation applications, such as surface and contour tracing~\cite{lin2022tactile} and bi-manual manipulation tasks involving lifting and re-orienting~\cite{lin2023bi}, have been demonstrated using control policies based on simulated deep reinforcement learning.  


\revised{\noindent\textbf{Model-based Methods.}}
We refer to \revised{``model-based methods'' 
as those employing simplified models to approximate real sensor performance.} \revised{These methods assume that a sensor's outputs follow a simple mathematical model based on the inputs. A small dataset is then used from the actual sensor to fine-tune or characterise this model. Compared to ``model-free'' machine learning approaches mentioned above, these assumed models are lightweight, requiring smaller training datasets and less computational time. When the model is well-designed, it also offers the advantage of better generalisability. 
Compared to physics-based methods, the models used in model-based approaches do not necessarily adhere to the physical principles of the sensors. As a result, they are not inherently accurate by design and may lack guaranteed generalisability. However, these simplified models have the potential to capture the complex behavior of physical components and can effectively address manufacturing variations or noise in the sensors.} 

While there are multiple potential advantages of model-based methods, the performance of those methods does hinge on the quality of the model design. For example, a capacitance-based \revf{tactile array sensor} was simulated by approximating the sensor reading blurriness with a Gaussian function and convolving it with the ground-truth contact geometry~\cite{pezzementi2010characterization}. This method relies on the assumption of the linear performance of the taxels but works adequately for low-resolution array sensors. A model-based method is used in ~\cite{si2022taxim} to simulate GelSight sensors by employing a polynomial look-up table to model the shape-to-color mapping of the sensor's optical performance, facilitating rapid generation of GelSight sensor images with different optical designs.

\subsection{Transferring simulated sensor output to the real world}
Connecting simulated sensors to real sensors is crucial, especially when using machine learning methods for tasks in which large-scale training datasets are required. There are two lines of thought in regard to this transference of data. One is that simulated sensors should produce readings closely resembling those of real sensors, enabling direct transfer of learned knowledge to real robots. The other suggests that provided the simulated sensor can capture sufficient physics to be useful, machine learning methods can then be used to bridge a substantive sim-to-real gap (see data-driven methods above). In the first case, researchers generate tactile images that closely resemble GelSight sensor images~\cite{gomes2019gelsight,gomes2021generation,gomes2023beyond,si2022taxim} for direct application of trained models to real robots, such as training a neural network to measure contact geometry~\cite{suresh2022shapemap}. However, an issue with this approach was that domain gaps, particularly in contact force and friction, persist in geometry-based simulations. 
\revised{One solution} is to approximate force-to-deformation correlations using heuristic models and by calibrating virtual friction coefficients based on real grasping experiments~\cite{si2022grasp}, which \revised{proved effective when the contact situation is not complicated.} 

More commonly, because systematic domain gaps exist between simulated and real sensors, there remains a need for methods to bridge these gaps. \revised{Standard methods to address the domain gap include transfer learning and generative modelling}. 
These methods have been applied to geometry-based tasks~\cite{jianu2022reducing} and tactile-based object manipulation~\cite{zhao2023skill}. \revised{Another approach is to add domain randomisation to the neural network input, to encourage the network to be robust \rev{to} the domain gaps with the sensor models.} 
Improved performance in shape recognition tasks can be achieved by augmenting simulated data with noise resembling sim-to-real differences~\cite{gomes2021generation}. Likewise, domain randomisation can be used for similar purposes~\cite{chen2024general}. Another approach is to employ hand-crafted processing methods, such as signal normalisation, to address domain gaps between simulated and real signals~\cite{xu2023efficient}.

\subsection{Challenges and Outlook}
One common dilemma faced in tactile sensor simulator research is the \textbf{trade-off between efficiency and accuracy}. Efficiency encompasses factors such as compute speed, dataset collection effort, and development complexity. All three methods covered above for simulating tactile sensors -- physical modeling, data-driven approaches and model-based methods -- have their own advantages and limitations. Currently, model-based methods appear to offer the most promising balance between accuracy and efficiency, although they rely on meticulously designed models. However, for complex sensors like those containing highly curved soft sensing mediums, suitable models may be unavailable. Another significant challenge in the simulation of tactile sensors is accurately \textbf{modeling contact forces and torques}, an aspect often overlooked in many simulators. Shear force simulation is particularly complex due to the intricate deformation of soft sensor skins.

\textbf{Are \revised{highly accurate} simulators necessary?} Some effective sim-to-real learning approaches combine basic physics simulations with well-designed domain transfer methods, enabling zero-shot transfer of machine learning models trained on simulated data to real-world scenarios \cite{zhao2023skill,zhao2024fots}. This provides an alternative solution to the efficiency-accuracy dilemma. Another approach involves simulating representative features of the sensor's output instead of the raw output itself~\cite{Khandate-RSS-23,qi2023general,yang2024anyrotate}. These representations could include ground truth contact geometry, or lower-dimensional vectors derived from raw data. In sim-to-real learning tasks, machine learning models can then use these representations as input rather than the raw sensor readings.

\section{Tactile Data Collection and Benchmarking}
\label{sec:benchmarking}


Benchmarking has, in recent years, been considered a key enabler in the advancement of robotics research. One example is the YCB object and model set~\cite{calli2015ycb}, which provides mesh models of a set of objects to researchers with which to evaluate grasping methods. Benchmarks have also been created to compare different reinforcement learning algorithms for continuous robot control tasks~\cite{mahmood2018benchmarking}, manipulation skill learning~\cite{james2020rlbench}, and deformable object manipulation~\cite{lin2021softgym}. However, it is more challenging to benchmark in robotics \textit{per se},  given the involvement of the robot body, complex motions like  grasping, the assorted hardware, and the wide variety of tasks, scenarios, and standards. Indeed, there are multiple factors that may affect the performance of the algorithms evaluated: robots of different degrees of freedom will exhibit varied behaviors; a certain arrangement of the table may be biased toward a certain algorithm; a metric could be more challenging to define in one task than in another.  

As the field of tactile robotics develops, the pursuit of advancing artificial touch requires comprehensive understanding, precise data collection, and robust benchmarking methodologies. This section considers the complexities of tactile data acquisition, examining the many aspects of current datasets for tactile robotics. \revised{This involves exploring diverse data sources and examining the associated challenges. As such, we provide an overview of the current state of, and potential future developments in, tactile data collection and benchmarking.}




\subsection{Trends in tactile dataset collection and benchmarking}
Recently, datasets of tactile data, as shown in Table~\ref{tab:benchmark}, have been collected to benchmark algorithms for tasks like fabric recognition~\cite{yuan2017connecting,luo2018vitac,li2019connecting,cao2024multimodal}. In most of the datasets, sensors of different types have been included to provide a multimodal input to the recognition algorithms.  GelFabric~\cite{yuan2017connecting} provides tactile videos from a GelSight sensor~\cite{yuan2017gelsight}, along with RGB images and depth maps. The fabrics are in a drape state when visual data are obtained, whereas they are placed flat on the table when tactile videos are collected. The ViTac~\cite{luo2018vitac} dataset also uses tactile videos of fabric interaction with fabrics placed flat on a table, but the RGB images capture the fabric textures, which provide a benchmark to compare algorithms for texture recognition~\cite{lee2019touching,cao2020spatio}. Other datasets such as VisGel~\cite{li2019connecting}, ObjectFolder~\cite{gao2023objectfolder} and PoseIt~\cite{kanitkar2022poseit} include everyday objects from the YCB object set~\cite{calli2015ycb}, and include different types of data, {\em e.g.}, impact sounds~\cite{gao2023objectfolder}, and force/torque sensor values~\cite{kanitkar2022poseit}. Some of these datasets use a robotic arm to move the sensor as it collects tactile data~\cite{kanitkar2022poseit,li2019connecting}, \revised{or a robot equipped with tactile sensors to capture rich tactile signals during the execution of exploratory procedures~\cite{chu_robotic_2015}}. Others use a hand-held tactile sensor, like ViTac~\cite{luo2018vitac}, Touch and Go~\cite{yang2022touch} and GelFabric~\cite{yuan2017connecting}.

Datasets have also been used to benchmark tactile sensor simulation methods ~\cite{gomes2021generation} or for reinforcement learning algorithms in which tactile sensing provides the input~\cite{pecyna2022visual}. \revised{When comparing simulation methods, it is important to establish a ground truth for benchmarking, which typically involves collecting tactile data from the real world or evaluating the robot’s performance in physical environments. To achieve this, a twin setup is often employed~\cite{gomes2021generation,gomes2023beyond}. In simulation, tactile sensors gather data from the simulated interactions, while in the real-world setup, the same tactile sensors are arranged using an identical configuration on the robot. This allows for a direct comparison, as the data from both the simulated and real environments can be aligned and analysed. In~\cite{gomes2021generation}, a set of test objects (3D-printed objects with primitive geometric shapes) were used as benchmarks to evaluate the performance of simulation methods. The accessibility of 3D-printing has facilitated further studies that use such objects for benchmarking tactile sensor simulations, as demonstrated in~\cite{wang2022tacto,si2022taxim,gomes2023beyond,chen2023tacchi,zhao2024fots}.}

\begin{table*}[htb]
    \centering
     \caption{Datasets for benchmarking tactile robotics} 
    \resizebox{\textwidth}{!}{%
    	\renewcommand{\arraystretch}{1}
	\centering
    \begin{tabular}{c||c|c|c}
    \hline
\textbf{\revised{Dataset}} &\textbf{Sensors}&\textbf{Data types}&\textbf{Data collection approach} \\
         \hline
         GelFabric~\cite{yuan2017connecting}& GelSight~\cite{yuan2017gelsight}, Canon T2i SLR camera and Kinect &Tactile videos, camera images, and depth images & Manual \\ 
         \hline
         ViTac~\cite{luo2018vitac}& GelSight~\cite{yuan2017gelsight} and Canon T2i SLR camera &Tactile videos and camera images & \revised{Teleoperated} \\ 
         \hline
         VisGel~\cite{li2019connecting}& GelSight~\cite{yuan2017gelsight}, a video camera &Tactile videos, camera videos & \revised{Robotic} \\ 
         \hline
         Gomes~\cite{gomes2021generation} &GelSight~\cite{yuan2017gelsight} and simulation  & Real and simulated tactile images
         & \revised{Robotic}
         \\ 
        \hline

        Touch and Go~\cite{yang2022touch} &GelSight~\cite{yuan2017gelsight} and camera  & Visual and tactile videos
         & \revised{Teleoperated}
         \\ 
        \hline
        ObjectFolder~\cite{gao2023objectfolder} &  3D Scanner,  microphone, GelSight~\cite{yuan2017gelsight} 
        & Visual images, impact sounds, and tactile readings & 
        \revised{Teleoperated}\&\revised{Robotic}\\
         \hline
        SoftSlidingGym~\cite{pecyna2022visual} & Camera, GelTip~\cite{gomes2020geltip} & Camera images, tactile readings& \revised{Robotic} \\
        \hline

        PoseIt~\cite{kanitkar2022poseit} & Camera, GelSight~\cite{yuan2017gelsight}, force/torque & RGB images, tactile images, \revised{force/torque} values& \revised{Robotic} \\
        \hline 
        PHAC-2~\cite{chu_robotic_2015} & BioTac~\cite{fishel2012sensing} & Tactile data, haptic adjectives & \revised{Robotic} \\
        \hline
    \end{tabular}}
    \vspace{-1em}
    \label{tab:benchmark}
\end{table*}

\subsection{Challenges and Outlook}
Compared to fields such as computer vision, it is significantly more challenging to benchmark in tactile robotics. The main reason for this is that tactile sensing is highly coupled with robotic action. In contrast, visual benchmarking tasks are usually passive, and therefore algorithms can be compared using a dataset previously collected by a camera or other visual sensors. 
\revised{Benchmarking in tactile robotics involves several additional considerations. For instance, the robot platforms used to move tactile sensors can introduce different capabilities depending on their degrees of freedom. Unlike well-defined visual tasks, such as classification, detection, and segmentation, tactile robotic tasks vary widely, typically including exploration, passive/active shape and texture recognition, or grasping and manipulation. The active nature of tactile robotics means that the policies employed by the robot agent also present new challenges in benchmarking.} 

\revised{To this end, we propose to leverage hierarchical benchmarking approaches in tactile robotics. For example, tactile sensors, representation learning, and policies could first be benchmarked separately, allowing for a more detailed assessment of each component's performance. Subsequently, these components could be evaluated as an integrated system to understand their combined effects. There is also a need to establish benchmarks for the comparison of different tactile sensors through well-designed experiments that consider factors such as sensor resolution, material properties, and response time. These steps would provide a more comprehensive framework with which to advance tactile robotics.}

Evaluation metrics are another challenge to consider when benchmarking tactile robotics. When comparing  simulation methods for vision-based tactile sensors, a set of vision-based evaluation metrics could be used to quantify the difference between the simulated tactile images and the real images~\cite{gomes2021generation}, {\em e.g.}, Structural Similarity (SSIM)~\cite{wang2004image}, Peak Signal-to-Noise Ratio (PSNR) and Mean Absolute Error (MAE). However, these vision-based evaluation metrics may not be suitable for assessing  differences in tactile readings. For example, the luminance and contrast elements in the SSIM metric are good indicators when measuring differences in visual images, but may not be suitable when comparing tactile images as they are indirectly related to contact force and movement of the tactile skin. Indeed, it raises the question: \textbf{what are the suitable evaluation metrics for benchmarking tactile robotics?} Broadly speaking, a good evaluation metric for tactile robotics should be task-relevant, sensitive to tactile measurement quality, robust across various conditions, repeatable, and computationally efficient. \rev{One underutilised source in robotics is psychophysics, which has developed various metrics for human tactile perception. For example, two-point discrimination and grating-orientation experiments have been used to probe artificial tactile spatial acuity~\cite{delhaye2016robo,pestell2022artificial-1}.}

\section{Tactile Data Interpretation}
\label{sec:perception}


In the previous sections, we discussed how to obtain data from tactile sensor equipped robots, either in the real world or in simulated environments. How should we make use of this data for robotic tasks and how should we interpret that data?

\subsection{Current methods for robotic tactile perception}
In recent years, several studies have attempted to predict physical stimulus information and object properties using the tactile sensors described in the previous sections.

\textbf{Force} is the most basic physical stimulus that researchers have attempted to predict in this way~\cite{chen2024deep,chen2025transforce}. Relevant studies can be grouped into a few distinct types: pressure and normal force perpendicular to the sensor-object contact, slip and shear force tangential to the normal force, and 6-axis force/torque combining all forces and their moments on the three axes. The distribution of normal pressure (force per unit area) across the robot fingertips can indicate the geometry of the object in contact. Shear force plays a pivotal role in various robotic tasks, such as object state estimation during grasping and surface friction prediction through tactile sensing. Force in tactile array sensors is directly calibrated during manufacturing, whereas vision-based tactile sensors infer force implicitly, typically via marker motion using learning-based~\cite{yuan2017gelsight} or inverse FEM methods~\cite{ifem20}.

\textbf{Slip} is an important tactile feature that is essential to prevent grasp failure. There are many ways to detect slip: measuring the normal force of contact and the shear force and comparing the friction coefficient \cite{melchiorri2000slip,reinecke2014experimental,veiga2018grip}, measuring a sudden increase in vibration \cite{howe1989sensing,yamada1994tactile,kondo2011development,su2015force}, measuring a sudden change in the shear strain of the sensor surface~\cite{james2020slip}, directly measuring the movement of objects on the sensor surface~\cite{dong2017improved}, and measuring the stretching or uneven displacement of the soft surface of tactile sensors \cite{yuan2015measurement,khamis2019novel}. Among those methods, vibration-based detection is a popular method since it provides a very strong signal even with a very simple sensor, such as an accelerometer attached to the skin. \re{In recent years, skin-stretch-based methods have also gained increasing popularity} because they are easy to apply to \re{vision-based sensors; they detect slip by tracking the planar motion of the markers on the surface and measuring the unevenness of the distribution.} Another approach is to use machine learning methods to detect slip in an end-to-end manner\cite{fujimoto2003Development,yan2022detection}. A recent work further explored this by introducing a supervised regressive model to predict marker motions for a tactile sensor without physical markers~\cite{ou2024marker}.

\textbf{Position, size and shape} are clearly fundamental features that have attracted interest in relation to robotic tactile perception in recent years. In optical tactile sensors like GelSight, photometric stereo~\cite{ackermann2015survey} has been used to estimate the surface normals of objects by observing them under different lighting conditions. Several variants of the GelSight, with differing LED colors, have therefore been used \cite{yuan2017gelsight,gomes2020geltip}. Hand-crafted features like image moments~\cite{li2013control,pezzementi2011tactile} or adapted image descriptors like Scale-Invariant Feature Transform (SIFT)~\cite{luo2015novel} have been widely used to predict the pose of the contact or the shape information of the in-contact object. Image descriptors can also help localise the contact on an object~\cite{luo2015localizing}, while deep neural networks have been used to represent the shape or geometric information of the in-contact object. The shapes of various 3D objects have been reconstructed using tactile sensing to refine the shape learned by vision~\cite{wang20183D}. This was improved to make use of advances in 3D vision to infer accurate 3D models of objects using the DIGIT optical tactile sensor~\cite{smith20203D, smith2021active}, which was then adapted to a purely tactile method using the TacTip with a DeepSDF network~\cite{comi2024touchsdf}.

The local pose of an object surface or edge relative to the tactile sensor is important for servo control, and both feature-based~\cite{li2013control,kappassov2020touch} and convolutional network-based~\cite{lepora2021pose} methods have been considered. This has recently been extended to incorporate tactile sensor shear~\cite{lloyd2024pose}, for example to track moving objects.


\textbf{Softness/hardness} is a physical property that can be inferred by touch by comparing the temporal change of the tactile signal, either in the form of normal contact force or the contact geometry. Intuitively, when pressing on a soft object with an increasing penetration depth, \re{the normal force may grow much slower compared to when pressing on a rigid object}, and the deformation of the soft object can be captured by tactile sensors and used to infer the hardness level. Researchers have used machine learning methods to model the correlation between hardness and tactile signals, 
\re{such as by controlling the pressing displacement and analysing the rate of force increase~\cite{yussof2008low,drimus2011classification,su_use_2012,wang2021tactual,he2024cherry}.} Alternatively, for sensors that can measure contact geometry, such as GelSight, this can be done by comparing the change of contact geometry and the normal force in a pressing sequence without precise control of the pressing trajectory~\cite{yuan2016estimating}. Following work used a neural network structure that combines CNNs and recurrent neural networks (RNNs) to model the hardness level from a labeled dataset~\cite{yuan2017shape,qian2019hardness,fang2022tactonet}. More recently, a biomimetic approach has been proposed to measure compliance based on the rate-of-change of tactile contact area~\cite{pagnanelli2024integrating}.

\textbf{Roughness and texture} have also been analysed using tactile sensors. \revised{A classic study used all modalities of a BioTac sensor for Bayesian exploration of textures~\cite{fishel2012bayesian}.} More recently, attention-based networks have been proposed to learn the salient features of fabric textures~\cite{cao2020spatio}. Similarly, a spiking neural network and a spiking graph neural network have been used  to learn the features of object textures~\cite{taunyazov2020fast,gu2020tactilesgnet} using flexible, event-driven, electronic skins.

\subsection{Challenges and Outlook}
\re{While most desired information from tactile signals has some connection with the physical model of the tactile sensor, extracting  quantitative measurements is more challenging due to the complex mechanics of the soft structure of the tactile sensors and the complicated nature of contact. Therefore, machine learning models can be the method of choice, often with end-to-end training: researchers collect corresponding tactile signals with precise labels and then train a machine learning model to predict the desired feature.}

\rev{The tactile sensing community has greatly benefitted from new machine learning technologies. In recent years, deep neural networks have become the most popular}
methods of extracting useful information from tactile data, in particular for recognising tactile features. \rev{Deep learning is particularly well suited for high-resolution tactile sensors that provide data with much higher information content than traditional sensors.}
However, \revised{the use of deep neural networks does bring some new challenges for tactile robotics, which we now discuss.} 

Firstly, deep neural network models are computationally expensive and challenging to deploy in local tactile processing units. In human perception, much of the processing, such as estimating contact events like contact force and slip, occurs in such local processing units. One remedy is to use neural architecture search techniques ({\em e.g.},~\cite{ren2021comprehensive}) to help reduce the size of the neural networks and improve inference time.

Secondly, deep learning methods are data hungry. For research problems like vision and audio processing, it is relatively easy to collect and annotate large amounts of data. However, in robotic tactile perception, large-scale datasets are costly. This is compounded by the fact that there are a variety of tasks and scenarios in robotic tactile perception. Future work could investigate how to use limited data while improving generalisation of the learning models. \revised{One potential approach is zero-shot learning, which transfers knowledge from touched objects to untouched ones, enabling their recognition. Recent studies have explored this by leveraging haptic attributes~\cite{abderrahmane2018haptic} or combining visual and semantic information~\cite{cao2024multimodal} to facilitate tactile zero-shot learning.}

Thirdly, sensors of different types have been used in robotic tactile perception and therefore different types of deep neural networks may be effective in processing the varied data. It would therefore also be interesting to build a foundation model for robotic tactile perception using different tactile sensors for different tactile perception tasks. \revised{Though much attention has been paid} to static contact in the past years, temporal information of contact events has not yet been fully investigated. \revised{Standard temporal neural networks, such as RNNs, have been used to process temporal data~\cite{cao2020spatio}. The human sense of touch generates spike trains that are transmitted to the brain for perception~\cite{yavari2020spike}, suggesting that spiking neural networks inspired by this biological process could be effective for processing tactile data~\cite{taunyazov2020event}.}


\section{Multimodal and Cross-modal Learning}
\label{sec:multicross}

\revised{Human interaction with the world is inherently multimodal, combining senses such as vision and touch. In contrast, robots often rely on isolated sensing modalities, with distinct hardware and algorithms designed for specific tasks and specific modalities.} The conventional division between disciplines like computer vision and tactile robotics persists, demanding specialised expertise to advance each research domain. However, envisioning future robots as embodied agents in human-centric environments calls for a paradigm shift. These robots will need to seamlessly integrate and leverage all available sensing modalities to optimise their performance in various tasks.

\subsection{Incorporating tactile sensing into multimodal and cross-modal human perception}

Vision and tactile sensing stand out as the principal modalities through which humans perceive and navigate the surrounding world. By coordinating eye and hand functions, we proficiently undertake complex tasks from recognition and exploration to intricate object manipulation. Vision, through which we are able to perceive the appearance, texture and shape of objects from a distance, complements touch, which is crucial in acquiring detailed information on texture, local shape, and other tactile properties through physical interaction. \revised{Our perceptual experience extends into the interplay between touching to see and seeing to feel~\cite{lee2019touching}.} 

Research in neuroscience and psychophysics has investigated the mutual interaction between vision and tactile sensing~\cite{spence2011crossmodal}. Notably, visual imagery has been identified as a contributor to tactile discrimination of orientation in individuals with normal vision~\cite{calvert2001crossmodal}. The human brain adeptly employs shared models of objects across diverse sensory modalities, facilitating the seamless transfer of knowledge between vision and tactile sensing~\cite{banati2000functional}. This intermodal information sharing proves particularly advantageous when one sense is impeded. An illustrative example is the human tendency to rely more on touch when confronted with textures containing intricate details that are challenging to discern visually~\cite{newell2005visual}. Such findings underscore the adaptive nature of sensory processing and highlight the complementary roles of vision and tactile sensing in various perceptual contexts.

\subsection{Current methods for Multi-/Cross-modal Learning}
\begin{figure}[t]
    \includegraphics[width=\linewidth]{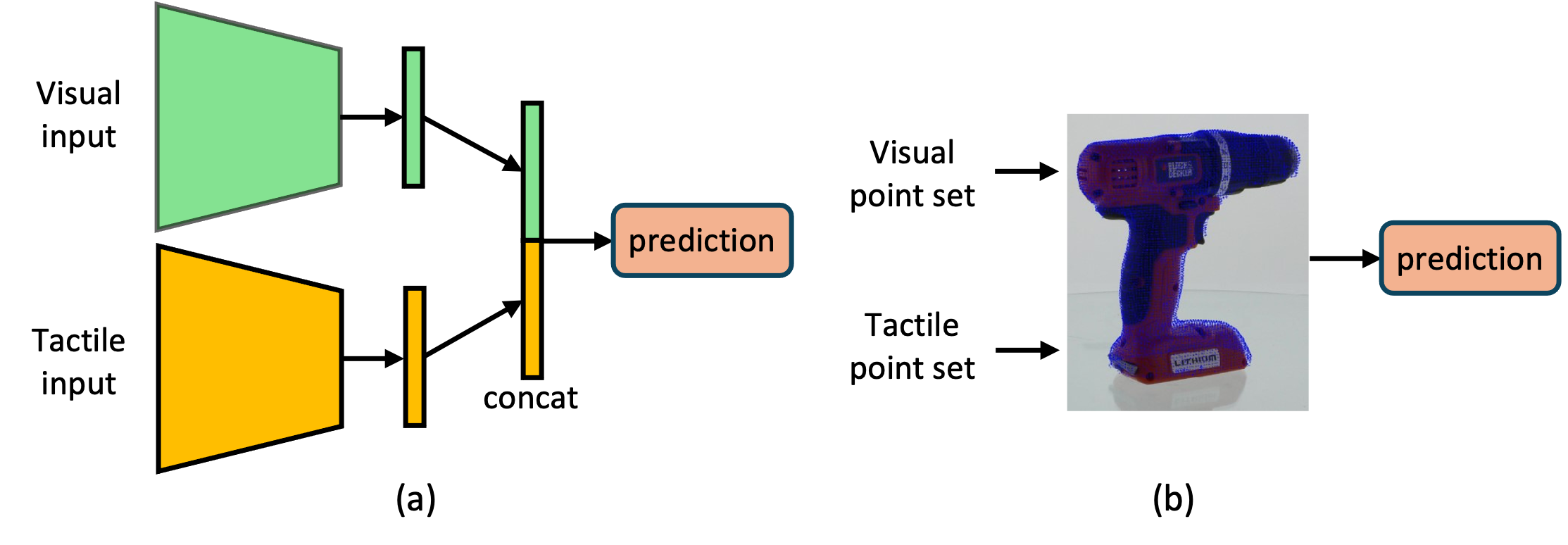}
        \vspace{-2em}
   \caption{\revf{Integration of vision and touch. (a) Vision and tactile features are extracted independently and fused via feature concatenation before being used for downstream prediction tasks like object classification or grasp success estimation. (b) Visual and tactile point clouds are aligned in a shared 3D space, forming a fused representation that enables joint reasoning for tasks such as object pose estimation or contact-aware manipulation.}}
 \label{fig:VisionTactileFusion}
 \vspace{-1em}
\end{figure}

Integrating vision and touch in robotics can be approached in a variety of ways, although these generally fall into two categories: either by combining features extracted from visual and tactile data, or by incorporating visual and tactile data into a representation of point clouds, as shown in Fig.~\ref{fig:VisionTactileFusion}.

\textbf{Feature-level fusion.} \revised{In early attempts to combine vision and tactile sensing dating back to the 1980s~\cite{allen1984surface}, tactile sensing played an ancillary role to vision, due to the lower resolution of tactile sensors at that time.} An innovative approach, exemplified in~\cite{bjorkman2013enhancing}, involves utilising tactile devices to confirm object-sensor contact. Visual features are first extracted to establish an initial hypothesis of object shape, and subsequently refined through the integration of tactile measurements. Alternatively, hand-designed features can be derived from tactile data and combined with visual features to form a multi-modal feature set. \revised{In works such as~\cite{bekiroglu2011learning}, image moments that statistically capture the spatial distribution properties of images were used to extract features from tactile data (both 2D and 3D). These moments were then integrated with visual and action features to create a comprehensive feature set tailored for grasping tasks.} Another study~\cite{guler2014s} addresses the identification of container contents through a dual approach: vision captures general container deformation, while tactile sensors record pressure distributions around contact regions during grasping. 

In a shift towards more sophisticated and integrated sensing methodologies, some studies have leveraged deep neural networks to extract features from both vision and haptic data~\cite{gao2016deep,pinto2016curious}. In ViTac, Maximum Covariance Analysis (DMCA) learned a joint latent space for sharing features through vision and tactile sensing in the task of fabric texture recognition~\cite{luo2018vitac}. Another approach is to use a
Siamese Neural Network (SNN)~\cite{chopra2005learning} to learn low-dimensional embedding vectors from both visual and tactile inputs~\cite{yuan2017connecting}. \revised{A late fusion approach can also combine vision and touch information by concatenating the feature vectors of visual and tactile data~\cite{lee2020making,tatiya2024mosaic} to form a multimodal representation, as shown in Fig.~\ref{fig:VisionTactileFusion}(a). The concatenated vector can then be fed into a fully-connected network that calculates the probability of producing a successful grasp~\cite{lee2020making}.}

\noindent\textbf{Point cloud-based methods.} Point cloud data play a crucial role in accurately estimating an object's pose. Given that both vision and tactile data can generate point clouds, their integration and alignment \revf{is particularly important for tasks such as pose estimation or manipulation, where joint reasoning over both modalities provides spatially grounded context}, as illustrated in Fig.~\ref{fig:VisionTactileFusion}(b). By initially deriving an object's geometry from vision data, an optimisation method, such as Gradient Descent or Levenberg Marquardt~\cite{marquardt1963algorithm}, can then be employed to obtain the transformation from the estimated object pose and the tactile sensor measurements~\cite{bimbo2013combining}. Another approach utilises the GelSight sensor to reconstruct a point cloud representation~\cite{izatt2017tracking}. \revised{All these methods showcase the efficacy of combining vision and tactile sensing for enhanced object pose estimation.}

\revised{Building on the integration of multiple sensory inputs, cross-modal learning has gained attention in recent years, using one sensory modality to improve robotic perception in another. For example, knowledge embedded in visual data can enhance tactile perception and vice versa.} One way is to pair vision and tactile samples to classify materials~\cite{kroemer2011learning}. Vision plays a pivotal role by offering a global overview of the scene, enabling tactile sensing to focus on extracting local information crucial for guiding the robot in planning its actions. The RGB image of the scene can also act as a reference map for tactile measurements~\cite{luo2015localizing}, which can facilitate the localisation of the robot within the map to enable an estimation of its pose. Alternatively, vision can identify potential poking areas and the robot then leverages this visual guidance to verify the location of the object~\cite{jiang2022shall}, enhancing its ability to grasp the object effectively. \revised{A cross-modal} framework has also been proposed for visuo-tactile object recognition, which involves  learning a subspace of vision and touch~\cite{falco2017cross}. 

There are also a few studies on \revised{cross-modal} data generation, which have been framed as cross-modal synthesis problems~\cite{radford2021learning}. Conditional generative networks can be used to generate new tactile data from visual data and vice versa~\cite{lee2019touching}, in the form of fabric textures~\cite{luo2018vitac}. Introducing a new conditional adversarial model can help the goal of reducing the scale discrepancy between vision and touch data~\cite{li2019connecting}: the former captures an entire visual scene in one go, whereas the latter
observes only a small region of an object at any given moment. Another method restyles the input image using a given touch signal as conditional information~\cite{yang2023generating}, {\em e.g.}, restyling a visual scene of rough rocks to match the smoother texture of brick.

\revised{The integration of touch into large multimodal datasets and models has gained attention in recent years. Studies such as Multiply~\cite{hong2024multiply}, TVL~\cite{futouch}, Octopi~\cite{yu2024octopi}, and Touch100k~\cite{cheng2024touch100k} all showcase the potential of large-scale multimodal datasets, typically combining vision, language, touch, and sometimes audio, to align sensory inputs across different modalities. These approaches either involve the creation of new datasets~\cite{hong2024multiply} or the augmentation of existing visual-language datasets with an additional modality, such as touch~\cite{cheng2024touch100k}. The alignment of tactile information with other sensory modalities can enhance perception and reasoning capabilities, enabling robots to better understand, interpret, and interact with their environments. However, these methods also present significant challenges due to the complexity of collecting multimodal data in interactive environments. Annotating tactile data presents a particular difficulty, as current technologies do not allow humans to experience recorded tactile sensations in the same way they can view photographs, watch videos, or listen to audio recordings. One possible solution is to align tactile embeddings with pre-trained image embeddings already associated with various other modalities through contrastive learning~\cite{yang2024binding}. This approach could help bridge the gap between tactile and other sensory modalities to facilitate more effective multimodal learning.}

\subsection{Challenges and Outlook}
Multimodal and cross-modal visual and tactile perception has progressed significantly in the past decade. However, several challenges have not been addressed, and we emphasise the need for increased attention to these areas in order to boost development of visual-tactile perception in robotic tasks.

\textbf{Beyond concatenating visual and tactile features.} The conventional approach of concatenating features learned from vision and tactile data has prevailed in much of the research~\cite{lee2020making,jagoda2024case}. However, this method treats visual and tactile features equally, overlooking their inherent correspondences. Vision captures color distribution and object appearance, aspects that are unobservable by touch, while touch captures local and detailed object information beyond visual reach. \revised{To better integrate these complementary modalities, correlation analysis has been introduced as a more effective bridging approach ~\cite{luo2018vitac}. 
Additionally, attention mechanisms~\cite{cao2020spatio,chen2022visuo} can be employed to dynamically weigh the contribution of each modality, allowing the model to focus on the most relevant features from vision and touch depending on the task. There is also work on visual-tactile models for task completion, using state-space models~\cite{chen2021multi} or a Reinforcement Learning framework~\cite{hansen2022visuotactile}.} Furthermore, semantic information can act as auxiliary data, providing a unified description for both visual and tactile observations. 

\textbf{The temporal dimension.} Current practices in multimodal and cross-modal perception feed data collected from visual and tactile sensors into neural networks for feature learning. However, these approaches often overlook the temporal correlation of vision and touch sensing, \rev{potentially crucial in tasks like grasping and manipulation.} Unlike static snapshots, the temporal relationship between visual and tactile data provides a dynamic understanding of the environment: vision can offer a global overview of the scene before any physical contact, while tactile sensing provides continuous, detailed feedback once interaction begins. Leveraging this temporal correlation between modalities could enhance the effectiveness of multimodal perception by providing a richer context and sequential understanding, going beyond what can be achieved with static data or point-cloud representations alone. Moreover, the necessity for high-speed sensing and rapid interpretation of tactile data becomes crucial when attempting to close the control loop in robotic systems. This is especially true when employing machine learning-powered sensors, as \rev{the ability to process tactile measurements in real time} directly affects the robot's responsiveness and adaptability.


\section{Active Tactile Perception}
\label{sec:active}

\subsection{Human active tactile perception}

Active tactile perception and active touch have been subject to evolving definitions over the past few decades~\cite{prescott_active_2011}. To understand the changes in terminology, we look back at previous definitions from a selection of key articles.

\subsubsection{Active touching}
\revised{Gibson's influential article} in 1962 on active touch~\cite{gibson_observations_1962} sheds light on the relation between bodily movements and the sense of touch. In this study, active touch, in which the perceiver initiates the tactile event, is distinguished from passive touch, in which an external force triggers the tactile sensation. Therefore, passive touch is not simply the absence of movement during tactile perception but rather the occurrence of a tactile event that is unanticipated. Gibson emphasised further that movements underlying active touch are intentional: the act of touching or feeling is a search for stimulation or, more precisely, an effort to obtain the kind of stimulation that facilitates perception of the object being touched. When an individual explores an object with their hands, the movements of the fingers are purposeful, as the body's sensory organs adjust and register the information. Active tactile perception seeks to help achieve an organism's goals by actively selecting and refining the sensations to access the best perceptual information.


\subsubsection{Active perception}
\revised{Another influential article, by Bajcsy~\cite{bajcsy_active_1988-1}, defined the term ``active perception''} as intentionally changing the sensor’s state parameters, based on sensing strategies for data acquisition, which depend on the current state of the data source, and the aim of the task. This aligns closely with Gibson’s notion of active touch, in which adjusting the sensor’s state parameters corresponds to moving or adjusting the sensor, data acquisition relates to sensing, and data interpretation mirrors perceiving. Taking an engineering perspective, Bajcsy emphasises that active perception is \revised{not simply feedback control applied to the state parameters of the sensor. It should rather be considered as a system in which feedback is delivered after complex processing of the sensory data, which includes} reasoning, decision making and control. As such, classical control theory is too limited for the implementation of active perception as it is designed for systems with simple relationships between sensory data and  variables to be controlled. 

Related definitions have been proposed for active vision. Aloimonos and colleagues defined active vision as occurring when an observer engages in an activity aimed at controlling the geometric parameters of the sensory apparatus~\cite{aloimonos_purposive_1990}. This has the purpose of manipulating the constraints that underlie the observed phenomena to improve the quality of the perceptual results. \revised{More recently, in 2018, these concepts from Aloimonos and Bajcsy were compiled, along  with related work on vision by Tsotsos, in an authoritative review article~\cite{bajcsy_revisiting_2018}.}

\subsubsection{Haptic exploration}
In Lederman and Klatzky’s \revised{influential} article on hand movements for haptic object recognition~\cite{lederman_hand_1987}, they propose that the hand, in conjunction with the brain, acts as an intelligent device by leveraging motor capabilities to greatly extend sensory functions. They introduce a taxonomy for purposive hand movements aimed at object apprehension, suggesting that specific exploratory procedures are associated with particular object dimensions. Like Gibson, they emphasise the purposive nature of hand movements in the recognition of objects, but go further in proposing that motor control is composed of a set of exploratory procedures that constitute those purposive movements. Although their work on haptic exploration and Bajcsy's definition of active perception were independent, they are related in that both were inspired by Gibson's original treatment of active touch. Haptic exploration can also be deemed as active from Bajcsy’s perspective, as in feedback control. Taking the exploratory procedure of contour following for example, a key pre-requisite is that the contour itself is perceived as controlling the finger's contact pathway while tracing.

\subsection{Methods for active robotic tactile perception}

Around the time that definitions of active perception and haptic exploration were being drawn in the human context, there were several initial proposals that robotic touch should be based on similar active principles~\cite{roberts_robot_1990,allen_mapping_1990}. Early examples of active perception with robotic tactile fingertips came soon after, examples being the motion control of a tactile fingertip for profile delineation of an unseen object~\cite{maekawa_development_1992} and a method to control the speed, and hence spatial filtering, of a tactile fingertip to measure surface roughness~\cite{shimojo_active_1993}. Another approach used active perception to compute contact localisation for grasping with a robot hand~\cite{kaneko1994contact}. A few years later, exploratory procedures were used to have some fingers of a robot hand grasp and manipulate the object while others feel the object surface~\cite{okamura_haptic_1997}. A survey of early work in this area was presented in the 2008 review on tactile sensing by Cutkosky, Howe and Provancher ~\cite{cutkosky2008force}. The 2010 review by Dahiya et al.~\cite{dahiya2009tactile} contrasts perception for action, such as grasp control and dexterous manipulation, with action for perception, such as haptic exploration and active perception.

In the 2010s, the principal attempts to implement exploratory active touch in robots were based on applying Bayesian inference to probabilistic models of tactile perception, inspired by the dominance of Bayesian models in psychology and neuroscience (the so-called ``Bayesian brain'' hypothesis~\cite{knill2004bayesian}). One approach to artificial exploratory touch used the spatial and vibrational modalities of a BioTac tactile sensor to implement a more human-like way of intelligently identifying tactile features. Exploratory haptic perception of textures was implemented by adjusting the normal force during a sliding motion to optimally distinguish the probability distribution of similar texture classes~\cite{fishel2012sensing}; likewise, \revised{tactile sensing} was used to control exploratory movements to characterise object compliance~\cite{su_use_2012}. These concepts of controlling the tactile sensor using Bayesian action and perception are related to perspectives from neuroscience on representing the world in the brain~\cite{loeb_bayesian_2014}. Meanwhile, another approach implemented robotic exploratory procedures such as \revised{squeezing} and sliding, using the BioTac, over a wide range of materials, to infer so-called ``haptic adjectives''~\cite{chu_robotic_2015}, which are the terms humans use when describing the feeling of objects. 

Independently of the approaches above, Bayesian methods have been applied to active touch for robust perception. Initially, these used a capacitive tactile sensor from the fingertip of an iCub robot, principally for perceiving the location and identity of objects~\cite{lepora_biomimetic_2016-4}. These methods were then extended to the haptic exploratory procedure of contour following around the edge of laminar surfaces~\cite{martinez-hernandez_active_2017-1}. A version of these contour-following methods was applied to the TacTip optical tactile sensor~\cite{lepora_exploratory_2017-1}, then reconsidered purely as tactile servo control using deep neural network models to perceive the contour orientation~\cite{lepora2019pixels}, generalising fully to contours sliding around complex 3D objects~\cite{lepora2021pose}. The latter methods no longer used a Bayesian approach, but just followed a contour without deciding autonomously how to explore an object. However, a Bayesian approach returned in later work with more complex tasks such as pushing and object tracking, in the form of a Bayesian filter to improve accuracy~\cite{lloyd2024pose}.

Other approaches have also combined tactile sensing with active perception to make decisions about object properties. A tactile-based framework for active object learning and discrimination was used with multimodal robotic skin~\cite{kaboli_tactile-based_2017}, then extended to active target object search in an unknown workspace~\cite{kaboli_tactile-based_2019}. Another approach, TANDEM, can jointly learn exploration and decision making with tactile sensors on planar objects~\cite{xu_tandem_2022}, which was extended to TANDEM3D for 3D object recognition~\cite{xu_tandem3D_2023}. In another study, active 3D shape reconstruction was demonstrated from vision and touch, using visual or tactile priors to guide the exploration~\cite{smith2021active}. \revised{Recently, a framework named Tactofind~\cite{pai2023tactofind} was introduced to utilise sparse tactile measurements from fingertip sensors on a dexterous hand to localise, identify, and grasp novel objects without visual feedback. Additionally, acoustic vibration sensing has been employed in SonicSense~\cite{liu2024sonicsense} to enhance tactile perception during object interaction.}

\subsection{Challenges and Outlook}

As things stand, active touch has found its place within robotics in purely perceptual tasks, such as texture classification or shape recognition. As active touch is fundamental to human tactile perception, which in turn is critical for human dexterity, there is clearly a broader question at hand about the role of active perception in robotics. The answer lies, perhaps, in Bajcsy's conception of active perception as an application of intelligent control theory that includes reasoning, decision making and control~\cite{bajcsy_active_1988-1}. 

\textbf{How to combine tactile control with active touch?} Active touch has usually been implemented alongside simple tactile control, which in most cases limits contact to basic taps, presses or sliding motions. These motions lack the dexterity that humans use to explore objects with touch. Guided motion under contact is a core feature of active touch, and therefore improved tactile control should also improve a robot's ability to actively interact with and explore objects.   

\revised{\textbf{How should tactile robots actively learn from experience?} A robot that is able to actively explore its environment is also a robot that can meet novel or unexpected situations. For example, during active touch a robot may meet new objects to learn new tactile sensations from. Lifelong and online learning are topical areas of machine learning that may give new insights into tactile perception and learning~\cite{dong2022lifelong}. Also, insights from how the brain implements action, perception and learning, as offered by the mathematical framework of active inference, may give a new perspective for robotics~\cite{lanillos2021active}.}




\section{Discussions and Conclusion}
\label{sec:conclusion}


Tactile robotics has found diverse applications across multiple fields. In \revised{MIS}~\cite{bandari2019tactile}, tactile information enhances precision and safety, while in robot manipulation, it can enable interaction with objects in unstructured and dynamic environments~\cite{chitta2011tactile,kollmitz2018whole}. Agriculture also benefits from tactile sensors in fruit harvesting and crop monitoring~\cite{ribeiro2020fruit,mandil2023tactile}. Tactile sensing feet can aid legged robots in navigating challenging terrains~\cite{zhang2021tactile}, while haptic displays translate tactile data into virtual experiences~\cite{cao2023vis2hap}, which can be made to feel real using touch interactive systems \revised{such as pseudo holograms}~\cite{christou2022pseudo}. Full-body tactile sensing in humanoid robots enhances human-robot interaction~\cite{argall2010survey}, enabling natural collaboration and assistive applications. Recently, multi-modal contact processing on humanoid robots with full-body tactile skin has demonstrated capabilities such as compliance, force control, and collision avoidance in dynamic environments. This gives clear evidence of the strides that are being made toward more sophisticated and interactive robotic systems~\cite{armleder2024tactile}.

Within the field of robotics more generally, tactile sensing is increasingly seen as a key frontier. Research is expanding and the outlook for this specific discipline is promising. While significant advances have been made in sensor technology, network architectures, simulations, tactile data collection and interpretation, multimodal and cross-modal learning, active touch perception and practical applications, many challenges still remain. Ongoing research efforts are, for example, looking into how to manage the vast amounts of tactile data and how to integrate heterogeneous sensor types.  Further challenges are \revised{likely to emerge with the increasing number of sensors} and associated electronics needed in some applications. Large skin and tactile sensor networks or robots connected by \revised{networked skin are prime examples.} In these areas, the emerging challenges are likely to be around distributed energy, computing, and communication hardware.

\revised{Due to space limitations, several important topics have not been covered in this paper. For instance, self-deformation and ego-vibrations, {\em i.e.}, tactile sensations caused by the robot’s own movement, are crucial aspects of tactile sensing, especially as robots become equipped with more sophisticated tactile systems. \re{The coupling between robot body motion and tactile sensing introduces challenges in distinguishing between self-caused and externally caused tactile sensations}. This issue is analogous to the challenge in computer vision of distinguishing camera motion from the movement of objects in the environment. Additionally, we have not discussed in detail the topic of sensor coverage: the proportion of the robot's body surface that can be instrumented with tactile sensors. This consideration depends further on whether soft tactile sensors are applied to rigid body parts or if the entire robot structure is soft. These topics will require further exploration as they are likely to play a significant role in the future development of robots with advanced tactile sensing capabilities.}

Looking ahead, research in tactile robotics will likely focus on enhancing sensor capabilities through novel designs and simulations, improving network communication, developing new methods for the processing of tactile data, \revised{facilitating the integration of multiple modalities}, and expanding the range of practical applications. Advances in machine learning will play a crucial role in interpreting and utilising tactile data for tasks such as object recognition, manipulation, and human-robot interaction. Tactile robotics could also gain by embracing  ongoing research in other areas such as printed batteries, printed electronics, and 5G/6G communication.

Overall, tactile robotics holds great potential for revolutionising various industries, from healthcare and manufacturing to agriculture and beyond. By continuing to innovate and collaborate across disciplines, researchers and engineers can unlock new possibilities for tactile sensing and pave the way for a future where robots interact with people, each other, and their environment with human-like dexterity and sensitivity.


\bibliographystyle{IEEEtran}
\bibliography{egibib,tactile_physiology}

\vfill

\end{document}